\documentclass{article}
\pdfoutput=1

    \PassOptionsToPackage{numbers, compress}{natbib}

    \usepackage[preprint]{neurips_2020}

\usepackage[utf8]{inputenc} 
\usepackage[T1]{fontenc}    
\usepackage{hyperref}       
\usepackage{url}            
\usepackage{booktabs}       
\usepackage{amsfonts}       
\usepackage{nicefrac}       
\usepackage{microtype}      
\usepackage{wrapfig}
\usepackage{algorithm}
\usepackage[noend]{algorithmic}

\title{Semantic Robustness of Models of Source Code}

\author{%
\vspace{-2em}
\hspace{-25pt}
Goutham Ramakrishnan\thanks{Equal Contribution}
\And \hspace{-30pt}
Jordan Henkel\footnotemark[1]
\And
Zi Wang
\AND
Aws Albarghouthi
\And
Somesh Jha
\And
Thomas Reps
\AND 
\vspace{-1.5em}
\\
University of Wisconsin--Madison \\
\texttt{\{gouthamr,jjhenkel,zw,aws,jha,reps\}@cs.wisc.edu}
}

\usepackage{subcaption}

\usepackage[tableposition=above,font=small]{caption}

\usepackage{amssymb}
\usepackage{amsmath}
\usepackage{amsfonts}

\usepackage{hyperref}
\usepackage[capitalize]{cleveref}

\hypersetup{
    colorlinks,
    linkcolor={red!50!black},
    citecolor={blue!50!black},
    urlcolor={blue!80!black}
}

\usepackage{listings}
\usepackage[scaled=0.80]{DejaVuSansMono}
\usepackage{colortbl}

\usepackage[dvipsnames,table]{xcolor}

\usepackage[warn]{textcomp}
\usepackage{pgfplots}
\usepackage{tikz}
 \usetikzlibrary{patterns}
\usetikzlibrary{arrows}
\usetikzlibrary{arrows.meta}
\usetikzlibrary{positioning}

\usepackage{microtype}
\usepackage{graphicx}
\usepackage{booktabs} 
\usepackage{filecontents}
\usepackage{multirow}
\usepackage{lipsum}  

\usepackage{lipsum}
\usepackage{csvsimple}
\usepackage{csquotes}

\usepackage{csquotes}
\usepackage{pifont}
\usepackage{tikz}
\usepackage{fp-basic}
\usepackage{wasysym}

\renewcommand{\leq}{\leqslant}

\makeatletter

\patchcmd{\NAT@test}{\else \NAT@nm}{\else \NAT@nmfmt{\NAT@nm}}{}{}

\DeclareRobustCommand\citepos
  {\begingroup
  \let\NAT@nmfmt\NAT@posfmt
  \NAT@swafalse\let\NAT@ctype\z@\NAT@partrue
  \@ifstar{\NAT@fulltrue\NAT@citetp}{\NAT@fullfalse\NAT@citetp}}

\let\NAT@orig@nmfmt\NAT@nmfmt
\def\NAT@posfmt#1{\NAT@orig@nmfmt{#1's}}

\makeatother

\crefname{section}{Sec.}{Secs.}
\definecolor{light-gray}{gray}{0.90}

\usetikzlibrary{arrows,backgrounds,decorations.markings,pgfplots.colorbrewer}
\usepgflibrary{shapes.multipart}
\usepgfplotslibrary{groupplots}
    
\tikzset{>=latex} 

\usepackage{enumitem}
\setlist[description]{leftmargin=1em,labelindent=1em} 

\tikzstyle{oper}=[rounded corners, draw=MidnightBlue, thick, minimum size = 6mm]
\tikzstyle{input}=[rounded corners, draw=Maroon, thick, minimum size = 6mm]
\tikzstyle{output}=[rounded corners, thick, draw=RoyalBlue, minimum size = 6mm]

\makeatletter
\lst@InstallKeywords k{attributes}{attributestyle}\slshape{attributestyle}{}ld
\makeatother

\lstdefinestyle{customc}{
  belowcaptionskip=1\baselineskip,
  breaklines=true,
  xleftmargin=\parindent,
  language=C,
  showstringspaces=false,
  basicstyle=\ttfamily,
  keywordstyle=\bfseries\ttfamily\color{MidnightBlue},
  commentstyle=\itshape\color{black},
  identifierstyle=\ttfamily\color{black},
  stringstyle=\itshape\color{blue},
  keywords={ map, flatMap, reduce, then, in, return,  for, if, else, reduceByKey, filter, partition},
moreattributes={let, where,int, def, Object}, 
attributestyle = \ttfamily\color{Mahogany}
}

\DeclareMathOperator*{\argmin}{{\rm argmin}}
\DeclareMathOperator*{\argmax}{{\rm argmax}}

\DeclareMathOperator{\calD}{{\mathcal D}}
\DeclareMathOperator{\calX}{{\mathcal X}}
\DeclareMathOperator{\calY}{{\mathcal Y}}
\DeclareMathOperator{\calH}{{\mathcal H}}
\DeclareMathOperator{\calR}{{\mathcal R}}

\DeclareMathOperator{\randTarg}{R}
\DeclareMathOperator{\gradTarg}{G}
\DeclareMathOperator{\attackSet}{\seqs}
\DeclareMathOperator{\noAttack}{\mathrm{Nor}}

\newcommand{\augTrain}{\mathrm{Tr}{\text -}{\mathrm{Aug}}}
\newcommand{\nrmTrain}{\mathrm{Tr}{\text -}{\mathrm{Nor}}}
\newcommand{\advTrain}[1]{\mathrm{Tr}{\text -}{#1}}
\newcommand{\testIdentity}{\noAttack^{\vphantom{1}}_{\;}}
\newcommand{\testRandDepthOne}{\attackSet^{1}_{\randTarg}}
\newcommand{\testGradDepthOne}{\attackSet^{1}_{\gradTarg}}
\newcommand{\testRandDepthFive}{\attackSet^{5}_{\randTarg}}
\newcommand{\testGradDepthFive}{\attackSet^{5}_{\gradTarg}}
\newcommand{\testGradDepthk}{\attackSet^{k}_{\gradTarg}}
\newcommand{\testRandDepthk}{\attackSet^{k}_{\randTarg}}

\newcommand{\tr}{t}
\newcommand{\trs}{\mathcal{T}}
\newcommand{\pr}{r}

\newcommand{\seq}{q}
\newcommand{\seqs}{\mathcal{Q}}

\renewcommand{\leq}{\leqslant}

\newtheorem{theorem}{Theorem}

\renewcommand{\circ}{{\scriptsize \color{black}\CIRCLE}}

\renewcommand\paragraph[1]{\noindent\textbf{#1}}
\begin{document}

\maketitle
\vspace{-1em}
\begin{abstract}
  Deep neural networks are vulnerable to adversarial examples---small input perturbations that result in incorrect predictions. 
  We study this problem for models of source code, where we want the network to be robust to source-code modifications that preserve code functionality.
  (1) We define a powerful adversary that can employ sequences of  parametric, semantics-preserving program transformations;
  (2) we show how to perform adversarial training to learn models robust to such adversaries;
  (3) we conduct an evaluation on different  languages and architectures, demonstrating significant quantitative gains in robustness.
  
\end{abstract}


\section{Introduction}\label{sec:intro}
While deep neural networks have been widely adopted in many areas of
computing, it has been repeatedly shown that they are vulnerable to
\emph{adversarial
  examples}~\cite{szegedy2013intriguing,biggio2013evasion,goodfellow2014explaining,papernot2017practical}:
small, seemingly innocuous perturbations to the input that lead to
incorrect predictions.
Adversarial examples raise  safety and security concerns, for example, in computer-vision models used in autonomous vehicles~\cite{eykholt2018robust,bhagoji2018practical} or for user authentication~\cite{sharif2016accessorize}.  
Significant progress has recently been made in identifying adversarial
examples and training models that are robust to such
examples.  However, the majority of the research has targeted
computer-vision tasks~\cite{carlini2017towards,madry2018towards,szegedy2013intriguing},
a continuous domain. 

\begin{wrapfigure}{r}{5cm}
\centering
\scriptsize
\vspace{-.25in}
\begin{lstlisting}[escapechar=!]
int !$\Box$!(Object target) 
  !\colorbox{light-gray}{System.out.println("Begin search");}!
  int i = 0;
  for (Object elem: this.elements)
      if (elem.equals(target))
          !\colorbox{light-gray}{System.out.println("Found");}!
          return i;
      i++;
  return -1;
\end{lstlisting}
\vspace{-1em}
\caption{code2seq~\cite{alon2018code2seq} correctly predicts function name: \texttt{indexOfTarget}. 
After the highlighted logging statements are added, it predicts \texttt{search}.}\label{fig:c2s}
\end{wrapfigure}
In this paper, we study the problem of robustness to adversarial examples in the discrete domain of deep neural networks for source code.
With the growing adoption of neural models for programming tasks, robustness is becoming an important property.
Why do we want robust models of code? 
There are many answers, ranging from usability to security.
Consider, for instance, a model that explains in English what a piece of code is doing---the \emph{code-captioning} task.
A developer using such a model to navigate a new code base should not receive completely different explanations for similar pieces of code.
For a concrete example, consider the behavior of the state-of-the-art code2seq model~\cite{alon2018code2seq} on the Java code in \cref{fig:c2s},
where the prediction changes after logging print statements are added.
Alternatively, imagine the security-critical setting of malware classification. We do not want a small modification to the malware's binary to cause the model to deem it safe. 

With images, the threat model involves \emph{small} changes that are imperceptible to a human.
With code, there is no analogous notion of a change imperceptible to a human.
Consequently, we consider attacks based on \emph{semantics-preserving transformations}.
Because the original program's semantics is preserved,
the program that results from the attack must have the same behavior as the original.

\paragraph{Most-related work.}
We are not the first to study this problem.
\citet{wang2019coset} demonstrated the drop in accuracy of deep models over source code when applying standard transformations and refactorings;
however, they did not propose defenses in their work.
Recently, \citet{yefet2019adversarial} developed a gradient-based attack that is specific to variable-name substitution or dead-code insertion, similar in spirit to attacks on natural-language models~\cite{ebrahimi2017hotflip},
which can efficiently estimate gradients for token substitution and insertion.
\citet{yefet2019adversarial} propose a defense based on outlier detection, but do not consider arbitrary program transformations or adversarial training.
\citet{zhang2020adversarial} use the Metropolis-Hastings algorithm to perform adversarial identifier renaming. 
However, they do not consider other program transformations, and their defense is akin to dataset augmentation with adversarial examples, instead of a robust optimization~\cite{madry2018towards}.
Parallel work~\cite{bielik2020adversarial}
is discussed in \cref{sec:related}.

Our aim in this paper is to consider the general setting of an adversary that can apply a sequence of source-code transformations.
Specifically, our goal is to answer the following question:
\begin{center}
    \emph{
    Can we train models that are robust to sequences of semantics-preserving transformations?
    }
\end{center}


\paragraph{$k$-adversaries \& $k$-robustness.}
We begin by defining the notion of a $k$\emph{-adversary},
one that can select a sequence of $k$ transformations from a prespecified set of semantics-preserving transformations $\trs$, and apply them to an input program.
The adversary succeeds if it manages to change the prediction of the neural network.
For example, a transformation may add dead code to a program, replace \emph{for} loops with \emph{while} loops, change variable names, replace one API call with an equivalent one, etc. 


The primary challenge in implementing a $k$-adversary
is the combinatorial search space of possible sequences of transformations.
Further, some transformations are parametric,
e.g., insert a print statement with string $s$, which
blows up the search space even further.
To implement a $k$-adversary in practice,
we exploit the insight that we can break
up the search into two pieces:
(1) \emph{enumerative search} through transformation sequences,
and (2) \emph{gradient-based optimization} to discover transformation parameters.
Specifically, given a sequence of transformations,
we \emph{partially apply} them to a given program,
without supplying parameters, resulting in a program \emph{sketch}~\cite{solar2006combinatorial}:
a program with unknown holes (tokens or AST leaves).
Then, gradient-based optimization, like that of \citet{yefet2019adversarial} or \citet{ebrahimi2017hotflip}, can be used to discover
a worst-case instantiation of the holes.

\paragraph{Adversarially training $k$-robust models.}
To train a $k$-robust model---one that is robust to $k$-adversaries---we adapt the robust-optimization objective of \citet{madry2018towards} to our setting:
%
Instead of computing the loss for a program $x$, 
we compute the worst-case loss resulting from a $k$-adversary.
While this approach is theoretically sound, it is hopelessly inefficient for two reasons.

First, a $k$-adversary is an expensive operation within the training loop, because it involves enumerating transformation sequences.
Thus, even for small values of $k$, the search space can make training impractically slow.
We demonstrate experimentally that training on a small value, $k=1$, results in models that are robust to larger values of $k$.
This phenomenon allows us to efficiently train adversarially without incurring a combinatorial explosion in the number of transformation sequences.

Second, 
program transformations are typically  defined as tree transformations over \emph{abstract syntax trees} (ASTs).
However, the neural network usually receives as input some other program format, some of which lose program structure or entire portions of the program---for example, tokens or subtokens of program text, or randomly sampled paths through the AST~\cite{alon2019code2vec,alon2018code2seq}.
Therefore, modeling an 
in-training adversary is very time consuming, because it
requires invoking external, non-differentiable program-analysis tools,
and converting back and forth between ASTs and training formats.

To work around training complexity, we  generate sequences of transformations offline and partially apply them to programs
to generate program sketches.
During training, we need only consider the generated program sketches, performing gradient-based attacks to search for optimal parameters of the transformations.
This approach allows us to avoid applying AST transformations during training.

\paragraph{Evaluation.}
We have developed an extensible framework for writing transformations
and performing adversarial training \cite{framework}.
%
We evaluate our approach on radically different programming languages---Java and Python---and architectures---using tokenized and AST program representations.  
%
Our results show that (1) adversarial training improves robustness
and
(2) training with a 1-adversary results in a model that is robust on larger values of $k$, enabling efficient training.
%

\paragraph{Contributions.} We summarize our contributions below:
\begin{itemize}
  \item
    We define \emph{$k$-transformation robustness} for source-code tasks:
    robustness to an adversary that is allowed $k$ transformations to an input program.
    We show how to implement a $k$-adversary by
    combining enumeration and gradient-based optimization over program sketches.
  \item
    We propose the first adversarial-training method for  neural models of code, adapting the robust-optimization objective of \citet{madry2018towards}.
    To efficiently model the adversary during training, we pre-generate
    program sketches by partially applying transformations.
  \item
    We build an extensible framework for adversarially training
    models of code. 
    We thoroughly evaluate our approach on a variety of
    datasets and architectures. Our results demonstrate improvements in
    robustness and the power of training with weak adversaries.
\end{itemize}



\section{Related Work}\label{sec:related}
In concurrent work,\footnote{A preprint of our work appeared earlier on arXiv than \cite{bielik2020adversarial}.} 
\citet{bielik2020adversarial} combine adversarial training with abstention and AST pruning to train robust models of code. 
There are a number of key differences with our work:
(1) We consider a richer space of transformations for the adversary,
including inserting parameterized dead-code.
(2) We use a strong gradient-based adversary
and program sketches for completing transformations, while they use a greedy search through the space of  transformations with a small number of candidates.
(3) Our adversarial training approach is more efficient, as it does not solve an expensive ILP problem to prune ASTs or train multiple models, 
but it is possible that we can incorporate their AST pruning in our framework.

\paragraph{Adversarial examples.}
In test-time attacks, an adversary perturbs an
example so that it is misclassified by a model (untargeted
attack) or the perturbed example is classified as an attacker-specified label (targeted)~\cite{athalye2018obfuscated,biggio2013evasion,ilyas2018black,chen2017zoo,carlini2017towards}. 
Initially,
test-time attacks were explored in the context of images. 
Our discrete domain is closer
to test-time attacks in natural language processing (NLP).  There are
several test-time attacks in NLP that consider discrete
transformations, such as substituting words or introducing
typos~\cite{Dimakis:NLP,mukund:NLP,ebrahimi2017hotflip,zhang2019adversarial,garg2020bae}. A key difference between our
domain and NLP is that in the case of programs one has to worry about
semantics---the program has to work even after transformations.

\paragraph{Deep learning for source code.}
Recent years have seen huge progress in deep learning for source-code tasks---see \citet{allamanis2018survey}.
Here, we considered (sub-)tokenizing the program, analogous to NLP,
and using a variant of recurrent neural networks. This idea has appeared in numerous papers,
e.g., the pioneering work of \citet{raychev2014code} for code completion.
We have also considered the AST paths encoding pioneered by \citet{alon2019code2vec,alon2018code2seq}.
Researchers have considered more structured networks, like graph neural networks~\cite{allamanis16convolutional} and  tree-LSTMs~\cite{zhao2018neural}. These would be interesting to consider for future experimentation in the context of adversarial training.
The task we evaluated on, code summarization, was first introduced by \citet{allamanis16convolutional}.



\section{Transformation Robustness}

We now formally define $k$-adversaries robustness and our robust-optimization objective. 


\paragraph{Learning problem.}
 We assume a data distribution $\calD$
over $\calX \times \calY$, where $\calX$ is the space of samples and
$\calY$ is the space of labels.  
As is standard, we wish to solve the
  following optimization problem:
\[
\argmin_{w \in \calH} \; \mathop{\mathbb{E}}_{(x,y) \sim \calD} \; L(w,x,y)
\]
where $\calH$ is the hypothesis space and $L$ is
the loss function. 
Once we have solved the optimization problem given above, we obtain a
$w^*$, which yields a function $F_{w^\star}: \calX \rightarrow \calY$.



\paragraph{Abstract syntax trees.}
We are interested in tasks where the sample space $\calX$ is that of programs
in some programming language.
We do not constrain the space of outputs $\calY$---it could be a finite set of labels for classification, a natural-language description, etc.

We view a program $x$ as an \emph{abstract syntax tree} (AST)---a standard data structure for representing programs.
The internal nodes of an AST typically represent program constructs,
such as while loops and conditionals, and the leaves of the tree
denote variable names, constants, fields, etc.
\cref{fig:trans} (left) shows the AST representation
of the code snippet \lstinline{if (x > 0) y = false}.
%

 
 
 


\paragraph{Transformations.}
A transformation $\tr$ of a program $x$ transforms it into another program $x'$.
Typically, $\tr$ is defined over ASTs, making it a tree-to-tree transformation.
Formally, we think of a transformation as a function $\tr :  \calR \times \calX \to \calX$,
where $\calR$ is a space of \emph{parameters} to the transformation.
For example, if $\tr$ changes the name of a variable, it may need 
to receive a new variable name as a parameter.

For our goal of semantic robustness, we will focus on \emph{semantics-preserving} transformations, i.e., ones that do not change the behavior of the program. 
Consider, e.g., the  transformation $\tr$
shown in \cref{fig:trans} (left), where we replace \texttt{x > 0} with \texttt{0 < x} (the transformed subtree is highlighted).
Note, however, that our approach is not tied to semantics-preserving transformations, and one could define transformations that, for example, introduce common typos and bugs that programmers make. 

\paragraph{Program sketches.}
It will be helpful for us to think of how to \emph{partially apply} a transformation $\tr$ to a program $x$, without supplying parameters.
Intuitively, this operation should result in a \emph{set of programs},
one for each possible parameter to $\tr$.
We assume that parameters only affect leaves
of the transformed AST. 
Therefore, when we partially apply a transformation $\tr$,
it results in a tree with unknown leaves.
Equivalently, we can think of such tree as a program \emph{sketch}:
a program with holes.
For example, if $\tr$ changes names of program variables to new given names,
applying $\tr$ partially to our running example
\lstinline{if (x > 0) y = false}
results in the program sketch
\begin{lstlisting}
     if ($\circ_1$ > 0) $\circ_2$ = false
\end{lstlisting}
where $\circ_i$ are distinct unknown variable names (holes) to be filled in the sketch.
The AST of this program sketch is shown in \cref{fig:trans} (right).

\begin{figure}[t!]
\small
\centering
    \begin{tikzpicture}
        \draw node at (1, 0) [input] (if) {\textbf{\texttt{if}}};

        \draw node at (0, -1) [input,fill={rgb:black,0.3;white,2}] (gt) {\texttt{>}};
        \draw node at (-.5, -2) [oper,fill={rgb:black,0.3;white,2}] (x) {\texttt{x}};
        \draw node at (.5, -2) [oper,fill={rgb:black,0.3;white,2}] (z) {\texttt{0}};

        \draw node at (2, -1) [input] (eq) {\texttt{=}};
        \draw node at (1.5, -2) [oper] (y) {\texttt{y}};
        \draw node at (2.5, -2) [oper] (f) {\texttt{false}};

        \draw[->,thick] (if) -- (gt);
        \draw[->,thick] (gt) -- (x);
        \draw[->,thick] (gt) -- (z);

        \draw[->,thick] (if) -- (eq);
        \draw[->,thick] (eq) -- (y);
        \draw[->,thick] (eq) -- (f);

        \draw[->, line width=.6mm] (2.7, -0.5) -- (3.5,-0.5) node[below] {$\tr$};

        \draw node at (1+4.5, 0) [input] (if1) {\textbf{\texttt{if}}};

        \draw node at (0+4.5, -1) [input,fill={rgb:black,0.3;white,2}] (gt1) {\texttt{<}};
        \draw node at (-.5+4.5, -2) [oper,fill={rgb:black,0.3;white,2}] (x1) {\texttt{0}};
        \draw node at (.5+4.5, -2) [oper,fill={rgb:black,0.3;white,2}] (z1) {\texttt{x}};

        \draw node at (2+4.5, -1) [input] (eq1) {\texttt{=}};
        \draw node at (1.5+4.5, -2) [oper] (y1) {\texttt{y}};
        \draw node at (2.5+4.5, -2) [oper] (f1) {\texttt{false}};

        \draw[->,thick] (if1) -- (gt1);
        \draw[->,thick] (gt1) -- (x1);
        \draw[->,thick] (gt1) -- (z1);

        \draw[->,thick] (if1) -- (eq1);
        \draw[->,thick] (eq1) -- (y1);
        \draw[->,thick] (eq1) -- (f1);
   \end{tikzpicture}
~~~\unskip\ \vrule\ ~~~
\begin{tikzpicture}
        \draw node at (1, 0) [input] (if) {\textbf{\texttt{if}}};

        \draw node at (0, -1) [input] (gt) {\texttt{>}};
        \draw node at (-.5, -2) [oper,fill={rgb:black,0.3;white,2}] (x) {$\circ_1$};
        \draw node at (.5, -2) [oper] (z) {\texttt{0}};

        \draw node at (2, -1) [input] (eq) {\texttt{=}};
        \draw node at (1.5, -2) [oper, fill={rgb:black,0.3;white,2}] (y) {$\circ_2$};
        \draw node at (2.5, -2) [oper] (f) {\texttt{false}};

        \draw[->,thick] (if) -- (gt);
        \draw[->,thick] (gt) -- (x);
        \draw[->,thick] (gt) -- (z);

        \draw[->,thick] (if) -- (eq);
        \draw[->,thick] (eq) -- (y);
        \draw[->,thick] (eq) -- (f);






   \end{tikzpicture}
   \caption{Left: Example AST transformation (no parameters). Right: AST of program sketch with two holes\vspace{-1.5em}}\label{fig:trans}
\end{figure}

\paragraph{$k$-adversary.}
Given a set of transformations $\trs$, a $k$\emph{-adversary}
 is an oracle that finds a sequence of transformations of size $k$ that maximizes the loss function.

We use $\seq$ to denote a sequence of transformations $\tr_1,\ldots,\tr_n$,
and their corresponding parameters $\pr_1,\ldots,\pr_n$,
where $\tr_i \in \trs$ and $\pr_i \in \calR$.
We use $\seq(x)$ to denote the program $\tr_n(\ldots\tr_2(\pr_2,\tr_1(\pr_1,x)))$,
i.e., the result of applying all transformations in $\seq$ to 
$x$.
Let $\seqs^k$ denote the set of all sequences of transformations and parameters of length  $k$.
Given a program $x$ with label $y$, the goal of the adversary is to transform $x$ to maximize the loss; formally, the adversary solves the following objective function:
\begin{align}
    \max_{ \seq
  \in \seqs^{k} } \; L(w,\seq(x),y) \label{eq:adv}
\end{align} 

\paragraph{Robust-optimization objective.}
Now that we have formally defined a $k$-adversary, we can solve a
\emph{robust-optimization problem}~\cite{ben2009robust} to learn a model that is robust to such attacks---we call this $k$-robustness.
Specifically, we solve the following problem:
\begin{align}
\argmin_{w \in \calH} \; \mathop{\mathbb{E}}_{(x,y) \sim \calD} \; \underbrace{\max_{ \seq
\in \seqs^{k} } \; L(w,\seq(x),y)}_{\text{objective of $k$-adversary}}
\label{eq:train-ideal}
\end{align}
Informally, instead of computing the loss of a pair $(x,y)$,
we consider the worst-case loss resulting from a $k$-adversary-generated
transformation to $x$.
Such robust-optimization objectives have been used
for training robust deep neural networks~\cite{madry2018towards}.
Our setting of source code, however, results in unique challenges, 
which we address next.

\section{Adversaries and Efficient Training}\label{sec:mismatch}
We now show how to practically implement a $k$-adversary and efficiently train $k$-robust models.

\subsection{Deploying $k$-adversaries}\label{sec:adv}
It is typically quite expensive to solve 
the optimization objective of a $k$-adversary (\cref{eq:adv}).
Even if we have $k=1$ and a single transformation,
the search space can be very large. 
For example, say we have a transformation
that changes names of function arguments.
For a function with $n$ arguments, the space of possible parameters
to the transformation is roughly $|\text{size of vocabulary}|^n$,
and vocabulary size is easily in the thousands for recent datasets.
Indeed, it can be easily shown that the \cref{eq:adv}
is PSPACE-hard, via a reduction from the PSPACE-complete problem of checking that the intersection of a set of automata is empty~\cite{kozen1977lower}.
(See proof in the Appendix.)
This is in contrast to the vision domain,
where finding adversarial examples is NP-complete~\cite{katz2017reluplex}.
 \begin{wrapfigure}{R}{0.55\textwidth}
  \vspace{-2em}
    \begin{minipage}{0.55\textwidth}
    \footnotesize
      \begin{algorithm}[H]
        
        \footnotesize
        \caption{$k$-adversary for program $x$ with label $y$}
        \begin{algorithmic}[1]
\STATE Let $x^\star = x$
 \FORALL{sequences $\tr_1, \ldots, \tr_k$}
    \STATE Let sketch $z[\cdot]$ be $\tr_k(\ldots\tr_2(\cdot,\tr_1(\cdot,x)))$ \label{line:trans}
    \STATE  \label{line:solve}
    Let $\pr = \argmax_{\pr\in\calR} L(w,z[\pr],y)$
    \STATE \textbf{if} {$L(w,z[\pr],y) > L(w,x^\star,y)$} \textbf{then} {$x^\star = z[\pr]$}

\ENDFOR        
    \STATE \textbf{return} $x^\star$

\end{algorithmic}\label{alg:adv}
      \end{algorithm}
    \end{minipage}
  \end{wrapfigure}
  
Clearly, a na\"ive enumeration approach is impractical for implementing a $k$-adversary.
And, unlike with robustness in the vision domain,
the space of AST transformations is not differentiable.
Nonetheless, we observe that we can break up the search space
into two pieces:
(1) an enumerative search over sequences of transformations
$\tr_1,\ldots, \tr_k$,
and (2) a gradient-based search over transformation
parameters, $\pr_1,\ldots,\pr_k$, which allows
us to efficiently traverse parameter space.

The full algorithm is shown in \cref{alg:adv}.
Given a sequence of transformations $\tr_1,\ldots,\tr_k$,
the algorithm partially applies it to the input program $x$.
This results in a program sketch $z[\cdot]$.
Recall that a sketch is a program with holes in the leaves of the AST,
and so we use $[\cdot]$ to denote a parameter that $z[\cdot]$ can take
to fill its holes; as with transformations, we use $\calR$
to denote the set of possible parameters to $z[\cdot]$.
At this point, we can use existing algorithms to discover a complete program $z[\pr]$ that maximizes the loss function (approximately).

For example, if the sketch is represented as a sequence of tokens
for an LSTM, the holes are simply missing tokens to be inserted in certain locations. As such, attacks from natural language processing, like HotFlip~\cite{ebrahimi2017hotflip}, can be used.
In our implementation, we use a version of a recent algorithm by \citet{yefet2019adversarial} that performs a gradient-based search 
for picking a token replacement. 
We first fill in the holes of the program sketch $z[\cdot]$ with temporary special tokens $s$. 
We replace the input embedding lookup layer with a differentiable tensor multiplication of one-hot inputs and an embedding matrix. 
This ensures differentiability up to the input layer $v$, allowing us to (1) take a gradient-ascent step in the direction that maximizes loss, and
(2) approximate the worst token replacement; formally, 
\[v' \leftarrow v + \eta \cdot \nabla_{v} L(w,z[s],y) \quad ; \quad r = \argmax \ v'\]
In other words, we pick the replacement $r$ as the token with the maximum value in the vector $v'$, obtained after a gradient-ascent step on the one-hot input $v$ corresponding to each special token in $z[s]$. 
We impose additional semantic constraints, e.g., in a sketch like  
\lstinline{if ($\circ_1$ > 0) $\circ_2$ = false}, we enforce that $\circ_1$ and $\circ_2$ receive different replacements.
See appendix for full details. 




\subsection{Adversarial training}
\label{ssec:adv_training}
To train a model that is robust to the $k$-adversary we defined in \cref{alg:adv},
we can solve the robust-optimization problem in \cref{eq:train-ideal},
where the inner maximization objective---the $k$-adversary's objective---is approximated using \cref{alg:adv}.
\begin{align}
\argmin_{w \in \calH} \; \mathop{\mathbb{E}}_{(x,y) \sim \calD} \; \underbrace{\max_{ \seq
\in \seqs^{k} } \; L(w,\seq(x),y)}_{\text{approximate using \cref{alg:adv}}}
\label{eq:train}
\end{align}

Practically, for every program $x$ in a mini-batch, we need to run \cref{alg:adv}
to compute a transformed program $x'$ exhibiting worst-case loss.
This approach is wildly inefficient during training for two reasons:
(1) the combinatorial search space of the $k$-adversary in \cref{alg:adv},
and (2) the mismatch between program formats for transformation
and for training.
We discuss both below.

\paragraph{Size of the search space.}
Given a set of transformations $\trs$,
\cref{alg:adv} runs for $|\trs|^k$ iterations,
a space that grows exponentially with $k$.
In practice, we observe that it suffices to train with a weak $k=1$ adversary,
and still be quite robust to stronger adversaries.
Therefore, the number of iterations of \cref{alg:adv} is restricted
to $|\trs|$ per training point $(x,y)$.
Analogous observations have been made in vision~\cite{zhang2019you,madry2018towards,wong2020fast} and NLP~\cite{DBLP:conf/emnlp/HuangSWDYGDK19}.

\paragraph{Representation mismatch.}
There is usually a mismatch between the program format needed for applying transformations and the program format needed for training a neural network. 
A program $x$ is an AST and the adversary's transformations are defined over ASTs,
but the neural network expects as input a different representation---%
for example, sequences of (sub)tokens, or as in a recent popular line of work~\cite{alon2019code2vec,alon2018code2seq}, a sampled set of paths from one leaf of the AST to another. 
%

Therefore, in \cref{alg:adv}, we have to translate back and forth between ASTs and their neural representation.
To be specific, line \ref{line:trans} of \cref{alg:adv} computes a sequence of transformations
over ASTs, resulting in a program sketch $z[\cdot]$.
Then, line \ref{line:solve} requires a neural representation of $z[\cdot]$
to solve the maximization problem.
This approach is expensive to employ during training: in every training step, we have to apply transformations using an external program-analysis tool and convert the transformed AST to  its neural representation.

We address this challenge as follows: 
%
%
To avoid calling program-analysis tools within the training loop, we pre-generate all possible program sketches considered by the adversary in \cref{alg:adv}. That is, for every program $x$ in the training set, 
before training, we generate the set of program sketches
$S_x = \{ z[\cdot] \ \mid\  z[\cdot] = \tr_k(\ldots\tr_2(\cdot,\tr_1(\cdot,x)))\}$
to avoid performing AST transformations during training.
As such, the robust-optimization objective in \cref{eq:train} reduces to the following:
\begin{align}
\argmin_{w \in \calH} \; \mathop{\mathbb{E}}_{(x,y) \sim \calD} \; \max_{ \pr\in\calR, z[\cdot] \in S_x } \; L(w,z[\pr],y)
\label{eq:train-eff}
\end{align}





\section{Experimental Evaluation}\label{sec:eval}

\paragraph{Research questions.}
We designed our evaluation to answer the following research questions:
\textbf{(Q1)} Does adversarial training improve robustness to semantics-preserving  transformations?
    \textbf{(Q2)} Does training with a $k$-adversary improve robustness against stronger adversaries (larger $k$)?

\begin{table}[t]
\caption{
Our suite of semantics-preserving transformations, along with the drops in F1 obtained by random ($\testRandDepthOne{}$) and gradient-based ($\testGradDepthOne{}$) adversaries \emph{per transformation}, for normally trained models on c2s/java-small. 
}
\label{table:transformations}
\centering
\scriptsize
\begin{tabular}{lcclcc}
\toprule
\multirow{3}{*}{Transform} & \multicolumn{1}{c}{seq2seq {\tiny (F1: 37.8)}} & \multicolumn{1}{c}{code2seq {\tiny (F1: 41.4)}} &
\multirow{3}{*}{Transform (Cont.)} & \multicolumn{1}{c}{seq2seq {\tiny (F1: 37.8)}} & \multicolumn{1}{c}{code2seq {\tiny (F1: 41.4)}} \\
\cmidrule(lr){2-2} \cmidrule(lr){3-3} \cmidrule(lr){5-5} \cmidrule(lr){6-6}
& \multicolumn{1}{c}{\tiny $\Delta$F1 ($\testRandDepthOne{}$ / $\testGradDepthOne{}$)} & \multicolumn{1}{c}{\tiny $\Delta$F1 ($\testRandDepthOne{}$ / $\testGradDepthOne{}$)} & 
& \multicolumn{1}{c}{\tiny $\Delta$F1 ($\testRandDepthOne{}$ / $\testGradDepthOne{}$)} & \multicolumn{1}{c}{\tiny $\Delta$F1 ($\testRandDepthOne{}$ / $\testGradDepthOne{}$)} \\
\midrule
\texttt{AddDeadCode} & 4.0  / 7.7 & 1.4  / \hphantom{1}2.9 & \texttt{RenameParameter} & 0.3  / 3.0 & 0.3  / 4.7 \\
\texttt{InsertPrintStatement} & 2.7  / 6.1 & 3.8  / 10.2 &  \texttt{ReplaceTrueFalse} & 0.0  / 0.7 & 0.2  / 0.5 \\
\texttt{RenameField} & 2.3  / 5.4 & 2.0  / \hphantom{1}2.0 &  \texttt{UnrollWhile} & 0.0  / 0.0 & 0.4  / 0.4 \\
\texttt{RenameLocalVariable} & 0.3  / 2.2 & 0.0  / \hphantom{1}2.5 &  \texttt{WrapTryCatch} & 2.5  / 9.4 & 1.4  / 7.8 \\
\bottomrule
\end{tabular}
\vspace{-1.5em}
\end{table}

\paragraph{Code-summarization task.}
We consider the task of automatic code summarization \citep{allamanis16convolutional}---the prediction of a method's name given its body---which is a non-trivial task for deep-learning models. The performance of models in this task is measured by the F1 score metric. 

We experimented with two model architectures: 
(i) a sequence-to-sequence BiLSTM model (seq2seq) and 
(ii) \citepos{alon2018code2seq} state-of-the-art code2seq model.
The seq2seq model takes sub-tokenized programs as inputs. 
We built upon the implementation of \citet{ibm2020pytorch}, and trained models for 10 epochs.
code2seq is a code-specific architecture that samples paths from the program AST in its encoder, and uses a decoder with attention for predicting output tokens.
We used the code2seq TensorFlow code~\cite{alon2018code2seq},
along with their Java and Python\footnote{
We modified the Python extractor
(making it similar to the Java one),
resulting in improved performance.} path extractors.
We train  code2seq models for 20 epochs.

\paragraph{Datasets.}
We conducted experiments on four datasets in two languages: 
Java, statically typed with types explicitly stated in the code, and Python, a dynamically typed scripting language.
The datasets originated from three different sources: (i) code2seq's java-small dataset (c2s/java-small) \citep{alon2018code2seq}, (ii) GitHub's CodeSearchNet Java and Python datasets (csn/java, csn/python) \citep{husain2019codesearchnet}, and (iii) SRI Lab's Py150k dataset (sri/py150) \citep{raychev2016probabilistic}. 
For computational tractability of adversarial training, we randomly subsample each dataset to train/validation/test sets of 150k/10k/20k each.


\paragraph{Program Transformations.}
We used two separate code-transformation pipelines: for Java, based on Spoon \citep{pawlak2015spoon}, and for Python, based on Astor \citep{berkerpeksag2020astor}.
These frameworks apply program transformations to generate program sketches, as described in \cref{ssec:adv_training}.
 %
We implemented a suite of 8 semantics-preserving transformations, listed in \cref{table:transformations}. All transformations except \texttt{UnrollWhiles} are parameterized; therefore, applying the transformations creates program sketches with \emph{holes}. The transformations are configured to produce only a \emph{single} hole (e.g., the \texttt{WrapTryCatch} transformation encloses the target method body in a \emph{single} try/catch statement and replaces the name of the variable holding the caught exception with a \emph{single} hole).
Refer to appendix for exact details.

\paragraph{Adversaries.}
We consider two variants of the adversary defined in \cref{alg:adv}, a weak and a strong adversary
that search through sequences of length $k$:

\begin{description}
\item[\emph{Random (weak) adversary}]$(\testRandDepthk{})$:  uses a random choice for the parameter $r$ in line~\ref{line:solve}, i.e.,
it randomly fills the hole in a sketch;
\item[\emph{Gradient-based (strong) adversary}]$(\testGradDepthk{})$:
 uses the gradient-based approach described in \cref{sec:adv} to fill sketch holes with parameters  that maximize loss in line \ref{line:solve}.
\end{description}

\cref{table:transformations} presents the drops in test performance from attacking normally trained seq2seq and code2seq models with each of the 8 transformations independently, i.e., using a 1-adversary that only has a single transformation in its set $\trs$.
We observe that the gradient-based attacks are significantly stronger than the random attacks 
(most cause 2$x$--15$x$ greater drops in F1).
Among the most effective transformations are \texttt{AddDeadCode}, \texttt{InsertPrintStatements}, and \texttt{WrapTryCatch}, which introduce new tokens in the program that confuse the predictor. 
Note that while the \texttt{UnrollWhiles} transformation has no effect in seq2seq's test performance by itself, it increases the strength of the other attacks when used in tandem with them. 

\begin{table*}[t]
\caption{
Evaluation of our approach and baselines across four datasets.
Numbers in brackets are the difference in adversarial F1 compared to
the normally trained model (higher is better).
}
\label{table:results}
\resizebox{\textwidth}{!}
{
\begin{tabular}{cccccccccc}
\toprule
\multirow{3}{*}{Model} &
\multirow{3}{*}{Training} &
\multicolumn{2}{c}{c2s/java-small} &
\multicolumn{2}{c}{csn/java} &
\multicolumn{2}{c}{csn/python} & 
\multicolumn{2}{c}{sri/py150} \\
\cmidrule(lr){3-4} \cmidrule(lr){5-6} \cmidrule(lr){7-8} \cmidrule(lr){9-10}
& & 
$\testIdentity{}$ & $\testGradDepthOne{}$ &
$\testIdentity{}$ & $\testGradDepthOne{}$ &
$\testIdentity{}$ & $\testGradDepthOne{}$ &
$\testIdentity{}$ & $\testGradDepthOne{}$  \\
\midrule
\begin{filecontents*}{data/table-seq2seq.csv}
modelName,trainingType,ctsJavaSmallId,ctsJavaSmallQOneR,ctsJavaSmallQOneG,ctsJavaSmallQFiveR,ctsJavaSmallQFiveG,csnJavaId,csnJavaQOneR,csnJavaQOneG,csnJavaQFiveR,csnJavaQFiveG,csnPythonId,csnPythonQOneR,csnPythonQOneG,csnPythonQFiveR,csnPythonQFiveG,sriPyId,sriPyQOneR,sriPyQOneG,sriPyQFiveR,sriPyQFiveG
seq2seq,$\nrmTrain{}$,37.8,29.5,23.3,18.8,11.5,32.3,23.5,17.2,15.9,9.8,28.9,21.7,16.2,15.2,10.4,34.3,29.3,22.0,25.2,16.5
,$\augTrain{}$,38.8,33.8 [+4.3],27.8 [+4.4],\textbf{31.7 [+12.9]},23.5 [+12.0],32.6,29.1 [+5.7],21.4 [+4.3],26.3 [+10.4],15.6 [+5.8],29.8,27.2 [+5.5],20.9 [+4.7],24.2 [+9.0],14.4 [+4.0],33.7,30.5 [+1.2],24.0 [+2.0],27.9 [+2.7],19.1 [+2.6]
,$\advTrain{\testRandDepthOne{}}$,40.0,33.6 [+4.1],27.5 [+4.2],30.4 [+11.6],21.7 [+10.2],38.1,34.0 [+10.6],30.1 [+12.9],30.6 [+14.7],25.1 [+15.3],36.6,33.3 [+11.6],29.8 [+13.5],30.8 [+15.5],25.3 [+14.9],41.8,37.8 [+8.5],33.4 [+11.4],34.7 [+9.5],28.1 [+11.6]
,$\advTrain{\testGradDepthOne{}}$,40.7,\textbf{34.8 [+5.2]},\textbf{32.0 [+8.7]},31.3 [+12.5],\textbf{26.3 [+14.8]},37.8,\textbf{34.3 [+10.8]},\textbf{32.8 [+15.6]},\textbf{32.1 [+16.1]},\textbf{29.7 [+19.9]},36.2,\textbf{33.5 [+11.8]},\textbf{32.2 [+16.0]},\textbf{31.6 [+16.4]},\textbf{29.4 [+19.0]},41.8,\textbf{38.1 [+8.8]},\textbf{37.1 [+15.1]},\textbf{35.4 [+10.2]},\textbf{33.4 [+16.8]}
\end{filecontents*}
\csvreader[
head to column names,
late after line=\\,
late after last line=\\\midrule
]{data/table-seq2seq.csv}{}%
{
  \ifthenelse{
    \equal{\modelName}{seq2seq}
  }{%
    \multirow{4}{*}{seq2seq}
  }{} &
  \trainingType & 
  \ctsJavaSmallId &
  \ctsJavaSmallQOneG &
  \csnJavaId &
  \csnJavaQOneG  &
  \csnPythonId &
  \csnPythonQOneG &
  \sriPyId &
  \sriPyQOneG
}%
\begin{filecontents*}{data/table-code2seq.csv}
modelName,trainingType,ctsJavaSmallId,ctsJavaSmallQOneR,ctsJavaSmallQOneG,ctsJavaSmallQFiveR,ctsJavaSmallQFiveG,csnJavaId,csnJavaQOneR,csnJavaQOneG,csnJavaQFiveR,csnJavaQFiveG,csnPythonId,csnPythonQOneR,csnPythonQOneG,csnPythonQFiveR,csnPythonQFiveG,sriPyId,sriPyQOneR,sriPyQOneG,sriPyQFiveR,sriPyQFiveG
code2seq,$\nrmTrain{}$,41.4,29.5,24.6,24.4,15.5,39.2,27.3,19.6,22.0,12.1,37.4,25.0,21.0,22.8,15.6,39.7,28.7,23.7,24.4,18.2
,$\augTrain{}$,42.3,29.4 [-0.1],23.4 [-1.2],23.4 [-1.0],14.5 [-0.9],39.4,26.9 [-0.4],19.7 [+0.1],20.5 [-1.5],11.8 [-0.3],36.8,25.5 [+0.5],20.8 [-0.2],22.8 [+0.0],15.8 [+0.2],39.7,29.1 [+0.4],24.4 [+0.7],24.8 [+0.4],17.8 [-0.4]
,$\advTrain{\testRandDepthOne{}}$,42.3,32.1 [+2.6],28.1 [+3.5],29.9 [+5.5],22.4 [+7.0],40.0,29.6 [+2.3],23.5 [+3.9],26.2 [+4.2],17.6 [+5.5],37.6,26.0 [+1.0],22.5 [+1.6],23.9 [+1.1],17.6 [+2.0],40.0,30.8 [+2.1],26.6 [+2.9],28.2 [+3.8],22.1 [+3.9]
,$\advTrain{\testGradDepthOne{}}$,43.6,\textbf{33.5 [+4.0]},\textbf{31.6 [+7.0]},\textbf{31.0 [+6.6]},\textbf{27.3 [+11.9]},40.6,\textbf{30.4 [+3.1]},\textbf{27.5 [+7.9]},\textbf{27.2 [+5.2]},\textbf{22.0 [+10.0]},37.3,\textbf{26.7 [+1.7]},\textbf{23.8 [+2.8]},\textbf{24.6 [+1.8]},\textbf{20.6 [+5.0]},40.2,\textbf{31.2 [+2.4]},\textbf{29.5 [+5.8]},\textbf{28.4 [+4.0]},\textbf{25.1 [+6.9]}
\end{filecontents*}
\csvreader[
head to column names,
late after line=\\,
late after last line=\\\bottomrule
]{data/table-code2seq.csv}{}%
{
  \ifthenelse{
    \equal{\modelName}{code2seq}
  }{%
    \multirow{4}{*}{code2seq}
  }{} &
 \trainingType & 
 \ctsJavaSmallId &
 \ctsJavaSmallQOneG &
 \csnJavaId &
 \csnJavaQOneG &
 \csnPythonId &
 \csnPythonQOneG &
 \sriPyId &
 \sriPyQOneG
}%
\end{tabular}
}
\vspace{-1.5em}
\end{table*}

\subsection{Robustness evaluation}
\paragraph{Our approach \& baselines.}
We will use $\advTrain{\testGradDepthOne{}}$
to denote models trained using our approach (\cref{eq:train-eff})
with the gradient-based 1-adversary.
We consider a series of progressively stronger baselines:
\begin{description}
\item[\emph{Normal training}] ($\nrmTrain{}$):
Models are trained using a standard (non-robust)
objective.
\item[\emph{Data augmentation}] ($\augTrain{}$):
Minimize the composite loss $\sum_i L(w, x_i,y_i) + L(w, x_i',y_i)$,
where, for each $x_i$, a program $x_i'$
is constructed
using a random transformation from $\seqs^1$.
\item[\emph{Random adversarial training}] ($\advTrain{\testRandDepthOne{}}$):
Similar to our adversarial-training method,
except it uses a random choice for parameters
$r$ (line~\ref{line:solve} of  \cref{alg:adv}), instead of one that maximizes loss.

\end{description}



\begin{figure}
\centering
\small
\begin{tikzpicture}
\pgfplotsset{
        cycle list/.define={my marks}{
            every mark/.append style={solid,fill=\pgfkeysvalueof{/pgfplots/mark list fill}},mark=*\\
            every mark/.append style={solid,fill=\pgfkeysvalueof{/pgfplots/mark list fill}},mark=square*\\
            every mark/.append style={solid,fill=\pgfkeysvalueof{/pgfplots/mark list fill}},mark=triangle*\\
            every mark/.append style={solid,fill=\pgfkeysvalueof{/pgfplots/mark list fill}},mark=diamond*\\
        },
    }
    \begin{groupplot}[
        cycle list/RdBu-4,
        mark list fill={.!85!white},
        cycle multiindex* list={
            RdBu-4
                \nextlist
            my marks
                \nextlist
            thick
                \nextlist
        },
        group style={
            group size=4 by 2,
            x descriptions at=edge bottom,
            y descriptions at=edge left,
            vertical sep=0pt,
            horizontal sep=0pt
        },
        xticklabels={
            $\testIdentity{}$,
            $\testRandDepthOne{}$,
            $\testRandDepthFive{}$,
            $\testGradDepthOne{}$,
            $\testGradDepthFive{}$
        },
        xtick=data,
        xmode=normal,
        ticklabel style = {font=\tiny{}},
        xtick pos=left,
        ytick pos=left,
        major grid style={line width=.1pt, draw=gray!50},
        minor grid style={line width=.1pt, draw=gray!25},
        grid=both,
        minor tick num=1,
        ymin=8,
        ymax=45
    ]
    
    \nextgroupplot[
        ymode=normal, title={c2s/java-small}, width=4.60cm,
        ylabel={\footnotesize{F1 (seq2seq)}}, ylabel near ticks
    ]
    \addplot table [sharp plot,x=x,y=c2sJavaNormal,col sep=comma] {data/graph-seq2seq.csv};
    \addplot table [sharp plot,x=x,y=c2sJavaAug,col sep=comma] {data/graph-seq2seq.csv};
    \addplot table [sharp plot,x=x,y=c2sJavaAdvRand,col sep=comma] {data/graph-seq2seq.csv};
    \addplot table [sharp plot,x=x,y=c2sJavaAdvGrad,col sep=comma] {data/graph-seq2seq.csv};

    \nextgroupplot[
        ymode=normal, title={csn/java}, width=4.60cm, ymajorticks=false
    ]
    \addplot table [sharp plot,x=x,y=csnJavaNormal,col sep=comma] {data/graph-seq2seq.csv};
    \addplot table [sharp plot,x=x,y=csnJavaAug,col sep=comma] {data/graph-seq2seq.csv};
    \addplot table [sharp plot,x=x,y=csnJavaAdvRand,col sep=comma] {data/graph-seq2seq.csv};
    \addplot table [sharp plot,x=x,y=csnJavaAdvGrad,col sep=comma] {data/graph-seq2seq.csv};
    
    \nextgroupplot[
        ymode=normal, title={csn/python}, width=4.60cm, ymajorticks=false
    ]
    \addplot table [sharp plot,x=x,y=csnPythonNormal,col sep=comma] {data/graph-seq2seq.csv};
    \addplot table [sharp plot,x=x,y=csnPythonAug,col sep=comma] {data/graph-seq2seq.csv};
    \addplot table [sharp plot,x=x,y=csnPythonAdvRand,col sep=comma] {data/graph-seq2seq.csv};
    \addplot table [sharp plot,x=x,y=csnPythonAdvGrad,col sep=comma] {data/graph-seq2seq.csv};

    \nextgroupplot[
        ymode=normal, title={sri/py150}, width=4.60cm, ymajorticks=false
    ]
    \addplot table [sharp plot,x=x,y=sriPythonNormal,col sep=comma] {data/graph-seq2seq.csv};
    \addplot table [sharp plot,x=x,y=sriPythonAug,col sep=comma] {data/graph-seq2seq.csv};
    \addplot table [sharp plot,x=x,y=sriPythonAdvRand,col sep=comma] {data/graph-seq2seq.csv};
    \addplot table [sharp plot,x=x,y=sriPythonAdvGrad,col sep=comma] {data/graph-seq2seq.csv};
    
    \nextgroupplot[
        ymode=normal, width=4.60cm,
        ylabel={\footnotesize{{F1 (code2seq)}}}, ylabel near ticks
    ]
    \addplot table [sharp plot,x=x,y=c2sJavaNormal,col sep=comma] {data/graph-code2seq.csv};
    \addplot table [sharp plot,x=x,y=c2sJavaAug,col sep=comma] {data/graph-code2seq.csv};
    \addplot table [sharp plot,x=x,y=c2sJavaAdvRand,col sep=comma] {data/graph-code2seq.csv};
    \addplot table [sharp plot,x=x,y=c2sJavaAdvGrad,col sep=comma] {data/graph-code2seq.csv};

    \nextgroupplot[
        ymode=normal, width=4.60cm, ymajorticks=false
    ]
    \addplot table [sharp plot,x=x,y=csnJavaNormal,col sep=comma] {data/graph-code2seq.csv};
    \addplot table [sharp plot,x=x,y=csnJavaAug,col sep=comma] {data/graph-code2seq.csv};
    \addplot table [sharp plot,x=x,y=csnJavaAdvRand,col sep=comma] {data/graph-code2seq.csv};
    \addplot table [sharp plot,x=x,y=csnJavaAdvGrad,col sep=comma] {data/graph-code2seq.csv};
    
    \nextgroupplot[
        ymode=normal, width=4.60cm, ymajorticks=false
    ]
    \addplot table [sharp plot,x=x,y=csnPythonNormal,col sep=comma] {data/graph-code2seq.csv};
    \addplot table [sharp plot,x=x,y=csnPythonAug,col sep=comma] {data/graph-code2seq.csv};
    \addplot table [sharp plot,x=x,y=csnPythonAdvRand,col sep=comma] {data/graph-code2seq.csv};
    \addplot table [sharp plot,x=x,y=csnPythonAdvGrad,col sep=comma] {data/graph-code2seq.csv};

    \nextgroupplot[
        ymode=normal, width=4.60cm, ymajorticks=false
    ]
    \addplot table [sharp plot,x=x,y=sriPythonNormal,col sep=comma] {data/graph-code2seq.csv};\label{p1};
    \addplot table [sharp plot,x=x,y=sriPythonAug,col sep=comma] {data/graph-code2seq.csv};\label{p2};
    \addplot table [sharp plot,x=x,y=sriPythonAdvRand,col sep=comma] {data/graph-code2seq.csv};\label{p3};
    \addplot table [sharp plot,x=x,y=sriPythonAdvGrad,col sep=comma] {data/graph-code2seq.csv};\label{p4};

    \end{groupplot}
\end{tikzpicture}
\caption{%
A comparison of normally trained ($\nrmTrain{}$, \ref{p1}), trained with dataset augmentation ($\augTrain{}$, \ref{p2}), adversarially trained with random parameters ($\advTrain{\testRandDepthOne{}}$, \ref{p3}), and adversarially trained ($\advTrain{\testGradDepthOne{}}$, \ref{p4}) models
on four adversaries, four datasets, and two model architectures (seq2seq and code2seq).
\vspace{-1.5em}
}
\label{fig:attack-graphs}
\end{figure}

\paragraph{Robustness against 1-adversary.}
 \cref{table:results}
 compares our approach and the three baselines
 on two metrics:
 (1) the \emph{test} F1 score ($\noAttack$),
 and 
 (2) the \emph{adversarial} F1 score (denoted $\testGradDepthOne{}$), where every test point is attacked with the gradient-based $k=1$ adversary
 with all the transformations in \cref{table:transformations}.
 
We make the following key observations:
 $\advTrain{\testGradDepthOne{}}$ models are significantly more robust to attack than all the baselines, especially for seq2seq. 
Consider the seq2seq models trained on the c2s/java-small dataset. 
While the normally trained model suffers a drop in F1 of 14 points, from 37.8 to 23.3, $\advTrain{\testGradDepthOne{}}$ only sees a drop of 8.7 points. 
Its final F1 of 32.0 is markedly higher than the other models.
On the csn/python dataset, the F1 of seq2seq $\nrmTrain{}$ drops 44\% from 28.9 to 16.2. 
$\advTrain{\testGradDepthOne{}}$ starts significantly higher at 36.2, and experiences a drop of only 4 points to achieve an F1 score of 32.2 under a $\testGradDepthOne{}$ attack. 
On average across datasets,  $\advTrain{\testGradDepthOne{}}$ achieves gains in F1 over $\nrmTrain{}$ of around 6 (27\%) and 14 (74\%) points for code2seq and seq2seq, respectively.  


\emph{Our results answer \textbf{Q1} in the affirmative: adversarial training increases robustness to attack via semantics-preserving program transformations}. 

\paragraph{Robustness against a 5-adversary.}
We also studied the robustness of the models against a much stronger adversary. 
In particular, we considered random and gradient-based  5-adversaries, denoted by $\testRandDepthFive{}$ and $\testGradDepthFive{}$, respectively. 
Because there are an intractable number ($8^5\approx32k$) number of transformation sequences of length 5, 
we use a randomly sampled set of sequences of length 5 (see Appendix for details).
In addition, we also evaluated the models against a random 1-adversary, denoted by $\testRandDepthOne{}$. 
The graphs shown in \cref{fig:attack-graphs} summarize our findings.

In general, the strength of the adversaries increases along the $x$-axis. 
Interestingly we observe that the $\testGradDepthOne{}$ attack is often stronger than the $\testRandDepthFive{}$ attack, reaffirming the strength of the gradient-based approach to completing sketches and maximizing loss. 
While the $\advTrain{\testRandDepthOne{}}$ and $\advTrain{\testGradDepthOne{}}$ models achieve comparable robustness to the random adversaries, $\advTrain{\testGradDepthOne{}}$ consistently outperforms $\advTrain{\testRandDepthOne{}}$ against the gradient-based adversaries. 
The performance of all models drops significantly against the $\testGradDepthFive{}$ attack, however the drops suffered by the $\advTrain{\testGradDepthOne{}}$ models are much less compared to the others, thus displaying remarkable robustness to the very strong attack. 
For example, for the $\testGradDepthFive{}$ attack on seq2seq models for csn/java, the F1 score of $\nrmTrain{}$ drops from 32.3 to 9.8, while the F1 of  $\advTrain{\testGradDepthOne{}}$ drops just 8.1 points from 37.8 to 29.7. 

\emph{Our results answer \textbf{Q2} in the affirmative:
adversarial training with a small $k$ can improve robustness against a stronger adversary---at least for the case of training with $k=1$ and attacking with $k=5$.}

\paragraph{seq2seq vs.\ code2seq.}
It is interesting to compare the robustness of the seq2seq and code2seq models. Although the baseline test F1 scores of code2seq models are much higher than the
respective seq2seq models, the models are just as vulnerable to attack. 
Whereas adversarial training makes the seq2seq model significantly more robust, the robustness gains seen in code2seq are much less. 
This phenomenon is especially evident in the Python datasets, where there is a gap of $\approx$16 F1 points between the $\nrmTrain{}$ and $\advTrain{\testGradDepthOne{}}$ models under $\testGradDepthOne{}$ attack  for seq2seq, but only 2.8--5.8 F1 points for code2seq (see \cref{table:results}).
The phenomenon can also be seen in the code2seq graphs in \cref{fig:attack-graphs}.

We conjecture that this difference arises because the code2seq model uses sampled program AST paths to make its prediction. 
A single program transformation may affect
several AST paths at once, thus making the code2seq model especially vulnerable to attack, even after adversarial training.

\section{Conclusion}
To the best of our knowledge, this work is the first to address adversarial training for source-code tasks via a robust-optimization objective. 
Our approach is general, in that it considers adversaries that perform arbitrary and parameterized sequences of transformations.
For future work, it would be interesting to study the effects of
adversarial training on other models, e.g., graph neural networks.
It would also be interesting to explore other source-code tasks and transformations, such as automated refactoring.


\bibliographystyle{plainnat}
\bibliography{biblio,jha-bib}

\clearpage

\title{
\emph{Appendix} \\
Semantic Robustness of Models of Source Code
}


\maketitle

\appendix 

\section{PSPACE-hardness of a $k$-adversary}

Let $D_1,\ldots,D_m$ be $m$ deterministic finite automata (DFAs) over a finite alphabet $\Sigma$.
Let $n$ be the size of the largest automaton,
measured as the number of states.
Let $\mathcal{L}(D_i)$ be the language defined by $D_i$.
The problem of deciding whether $\cap_i \mathcal{L}(D_i)$ is empty is PSPACE-complete~\cite{kozen1977lower}.

Construct a program $x$ with $m$ variables $v_1,\ldots,v_m$
with assignment statements that assign each $v_i$ the initial state of the DFA $D_i$. 
Our reduction uses just
one transformation $\tr$ parameterized by the alphabet $\Sigma$. $\tr(\alpha,x)$ (where $\alpha \in \Sigma$)
transforms the program $x$ by updating the rhs of the assignment statements to the next states of all automata, 
i.e., $\tr$ simulates a single step of all the automata $D_i$.
We also include another transformation that is the identity function, i.e., $\tr(\cdot, x) = x$.
We assume a special stuck state.
Now we construct a $0-1$ loss function as follows:
$L(x) = 1$ iff the assignment statements in $x$ correspond to accepting states in all DFAs $D_i$. 

Note that the intersection is empty iff there is no string of length at most $n^m$ that is accepted by all automata.
(E.g., for $m=1$, if $D_1$ is non-empty, there has to be a string of length $\leq n$, the number of states.)
Therefore, it is
easy to see that $\max_{\seq \in \seqs^k} L(x)$ is $1$ iff $\cap_{i=1}^n L(D_i)$ is not empty,
where $k = n^m$.

Since PSPACE is closed under complement, we obtain the following theorem.

\begin{theorem}
The decision problem corresponding to maximizing the loss \[\max_{\seq \in \seqs^k} l(q(x))\] is  PSPACE-hard.
\end{theorem}




\section{Data Pipeline}
\subsection{Overview}

\begin{figure}
    \centering
    \includegraphics[width=\textwidth]{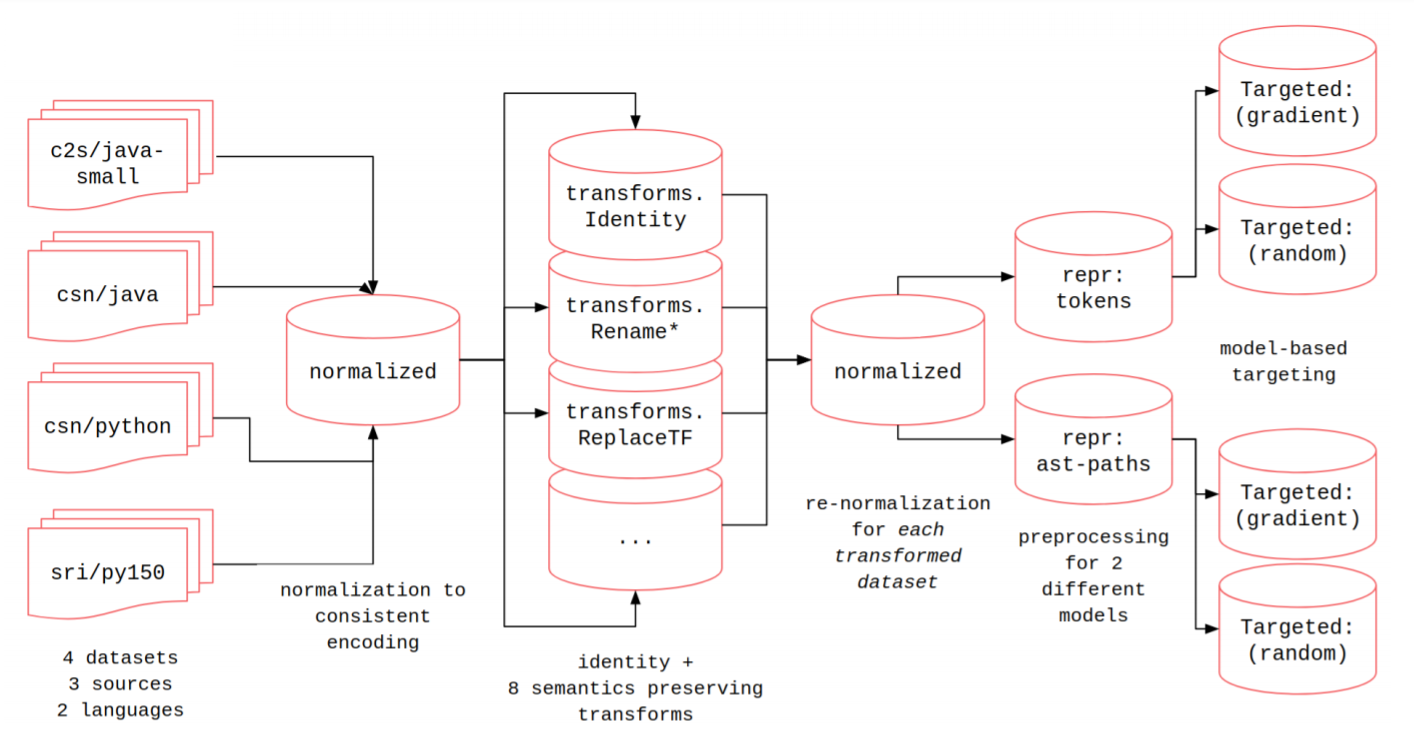}
    \caption{An overview of 
    our framework, which enables our experimental evaluation and bootstraps efforts towards training more robust models of source code. 
    With its plug-and-play architecture it can, in the future, be extended with more models, data sources, and program transforms.}
    \label{fig:averloc}
\end{figure}

\Cref{fig:averloc} shows an overview of the framework we built to support our experiments and bootstrap future efforts in the space of semantically robust models of source code. In general, the framework supports arbitrary input datasets. For each input dataset, a normalization procedure must be supplied to translate the input data into a consistent encoding. Once data is normalized, the rest of the  framework can be utilized to (i) apply program transformations to arbitrary depth, (ii) train models (normally), (iii) apply either gradient or random targeting to the holes in transformed programs (for seq2seq and code2seq), (iv) adversarially train models, and (v) adversarially attack and evaluate the trained models.
The source code is available anonymously~\cite{framework}.

\subsection{Details of Program Transformations}

We employ a total of eight semantics preserving transforms. Each program transform produces, as output, a \textit{program sketch} (that is, a program with one or more \textit{holes}). To turn a \textit{program sketch} into an complete program, each hole must be replaced with a valid token. For each of the eight transforms we implemented, valid hole replacements take the form of either variable names or string literals. Detailed descriptions of our eight semantics preserving transforms are given below:

\begin{enumerate}
    \item \textbf{\texttt{AddDeadCode:}} a single statement of the form \texttt{if (false) \{ int <HOLE> = 0; \}} is appended to the beginning or end of the target program. The insertion location (either beginning, or end) is chosen at random.
    \item \textbf{\texttt{RenameLocalVariables:}} a single, randomly selected, local variable declared in the target program has its name replaced by a hole.
    \item \textbf{\texttt{RenameParmeters:}} a single, randomly selected, parameter in the target program has its name replaced by a hole.
    \item \textbf{\texttt{RenameFields:}} a single, randomly selected, referenced field (\texttt{this.field} in Java, or \texttt{self.field} in Python) has its name replaced by a hole.
    \item \textbf{\texttt{ReplaceTrueFalse:}} a single, randomly selected, boolean literal is replaced by an equivalent expression containing a single hole (e.g., \texttt{("<HOLE>" == "<HOLE>")} to replace \texttt{true}).
    \item \textbf{\texttt{UnrollWhiles:}} a single, randomly selected, \texttt{while} loop in the target program has its loop body unrolled exactly one step. No holes are created by this transform.
    \item \textbf{\texttt{WrapTryCatch:}} the target program is wrapped by a single \texttt{try \{ ... \} catch (...) \{ ... \}} statement. The catch statement passes along the caught exception. A hole is used in the place of the name of the caught exception variable (e.g., \texttt{catch (Exception <HOLE>)}).
    \item \textbf{\texttt{InsertPrintStatements:}} A single \texttt{System.out.println("<HOLE>")}, in Java, and \texttt{print('<HOLE>')} in Python, is appended to the beginning or end of the target program. The insertion location (either beginning, or end) is chosen at random. 
\end{enumerate}

\section{Experimental Details}

\subsection{Datasets}
We conduct our experiments on four datasets in two different languages (java, python): c2s/java-small, csn/java, csn/python and sri/py150. 
Each dataset contains over 0.5M data points (method body--method name pairs). 
The original sizes of the datasets are shown in \cref{tab:datasets_table}.

Adversarial training in our domain of source code is an expensive process: we have to pre-generate several transformed versions of each data point as program sketches, and then repeatedly run the gradient-directed attack to fill in the holes in the sketches. 
Running extensive experiments across the 4 datasets and 2 models (seq2seq, code2seq) was computationally intractable, both in terms of time and space. 
Thus, we randomly subsample the four datasets to have train/validation/test sets of 150k/10k/20k each. 
The datasets remain sizeable, and thus we find that this only has a minimal effect on model performance.

\subsection{Models}
\paragraph{seq2seq}
The seq2seq models were given sub-tokenized programs as input, which were obtained by splitting up camel and snake-case, and other minor preprocessing.  
We trained 2-layer BiLSTM models, with 512 units in the encoder and decoder, with embedding sizes of 512.

\paragraph{code2seq}
code2seq is the current state-of-the-art model for the task of code summarization.
We built upon the implementation from the authors of code2seq\footnote{\url{https://github.com/tech-srl/code2seq}}, and use the original model parameters. 

For both seq2seq and code2seq, we use input and output vocabularies of size 15k and 5k respectively.
All models were trained and evaluated using NVIDIA GPUs (GeForce RTX 2080 Ti and Tesla V100).

\begin{table}[t]
    \centering
    \begin{tabular}{lrrr}
        \toprule
        Dataset & Train & Validation & Test \\
        \midrule
        c2s/java-small & 500.3k & 26.3k & 38.9k \\
        csn/java & 454.3k & 15.3k & 26.9k \\
        sri/py150 & 693.7k & 81.7k & 86.0k \\
        csn/python & 408.2k & 22.8k & 21.9k \\
        \bottomrule
    \end{tabular}
    \caption{The original sizes of the datasets}
    \label{tab:datasets_table}
\end{table}2

\subsection{Adversarial Training}
Adversarial training is known to improve robustness to attack at the cost of degraded performance on clean data~\cite{tsipras2018robustness}.
To maintain performance on clean data, we train with the following composite loss: 
\[
\sum_i \lambda \cdot L(x_i,y_i) + (1-\lambda) \cdot L_{adv}(x_i,y_i)
\]
where $L$ is the normal training loss and $L_{adv}$ is the loss from the robust optimization objective described in Sec. 4. 
The $\lambda$ hyperparameter controls the trade-off between performance and robustness, we picked $\lambda=0.4$ in our experiments after a grid search. 

The robust optimization objective for adversarial training requires choosing the worst transformation for each data point in each mini-batch of training. 
While this can be efficiently implemented in computer vision using PGD, it is very expensive to do so in our setting. 
For tractability of training, we do the following: 
\begin{itemize}
    \item We apply the gradient attack periodically during training to generate new token replacements for the program sketches. 
    For seq2seq we do this after every epoch, and for code2seq we do this after every two epochs. 
    \item Instead of picking the worst transformation for each individual point in a mini-batch, we pick the transformation that does the worst for all the points on the whole (i.e. highest loss over the batch). 
\end{itemize}

\subsection{Gradient Attack}
We adapt the attack for code models from \citet{yefet2019adversarial}, to find replacements for the holes in the program sketches. 
We represent \emph{holes} in the program sketches by special tokens. 
A sketch obtained after applying more than one transform would contain multiple variants of the special tokens, as the \emph{holes} must be filled in independently. 
The summary of the attack on seq2seq is described below:
\begin{itemize}
    \item During the training of the model, the special tokens are added to the input vocabulary.
    The embedding layer in the model encoder is replaced by an embedding matrix. 
    The input tensor of shape (batch\_size, max\_seq\_len) is converted to its one-hot representation $v$, a tensor of shape (batch\_size, max\_seq\_len,$\mid$source\_vocab$\mid$).  
    The embedding look-up is replaced by a multiplication between $v$ and the embedding matrix, the rest of the encoder remains the same.
    This step is necessary for differentiability upto the token-level input representation.
    \item For simplicity, we implement the attack to depth=1 and width=1. After one forward, we calculate the loss and backpropagate its gradient to the $v$ layer. 
    We take a gradient-ascent step on $v$, to yield $v'$. We approximate the worst token replacement for each special token by averaging gradients over each occurrence in the input and choose the token corresponding to the maximum. 
    We impose an additional semantic constraint that the token replacements made for different special tokens are distinct. 
\end{itemize}
In code2seq, the subtoken embedding layer is replaced by the tensor multiplication.

\begin{table}[]
    \centering
    \begin{tabular}{ccc}
        \toprule
        Number of Samples & Adversarial F1 & $\Delta$F1 (Baseline: $37.8$) \\
        \midrule
        10 & 29.7 & -8.1 \\
        20 & 28.5 & -9.3 \\
        30 & 27.9 & -9.9\\
        40 & 27.6 & -10.2 \\
        50 & 27.4 & -10.4 \\
        \bottomrule
    \end{tabular}
    \caption{An exploration of increasing sample counts for  $\testGradDepthFive$ attacks, on the $\advTrain{\testGradDepthOne}$ model on the csn/java test set. Note that the attack strength saturates very quickly.}
    \label{tab:attack-saturation}
\end{table}

\subsection{Implementing a tractable 5-Adversary}
In Sec. 5.1, we evaluate the robustness of the trained models on the 5-adversary $\testGradDepthFive$, i.e. an adversary that performs 5 semantics preserving transformations on the input. 
With our suite of 8 transforms, it results in a total of $8^5=32768$ possible sequences against which we need to evaluate each test data point. 
This is computationally intractable for two reasons:
\begin{itemize}
    \item For each of the 20k test points in each dataset, we choose the sequence of transforms which results in the greatest loss as the \emph{attack}. 
    To evaluate against each of the $8^5$ transforms, it would require millions of forward passes through the model.
    \item Each sequence of transforms is generated in two steps: (i) application of transformations to generate program sketches, and (ii) gradient attack to fill in the holes of the sketches. 
    Both these steps are expensive and time consuming. 
    In particular, generating 10 transforms for 20k data points requires 10 minutes for step (i) and 120 minutes for step (ii) for the code2seq model (seq2seq is somewhat faster, but still very slow). In addition, after generating program sketches and using a gradient attack to fill holes, we still must perform adversarial evaluation which, for code2seq on 200k data points (20k points $\times$ 10 transformed variants per point), takes an additional 60 minutes.
\end{itemize}
Due to the above reasons, we choose to implement the 5-adversary by randomly sampling 10 sequences of transformations of length 5. 
We investigate the effect of increasing the number of transforms sampled, and we find that the strength of the attack saturates. 
This is seen in \Cref{tab:attack-saturation} for the $\advTrain{\testGradDepthOne}$ model on the csn/java dataset. Because of this prominent saturation, we chose to use only 10 samples to keep total evaluation time to less than 24 hours (using 4 V100 GPUs).

\section{Full Robustness Evaluation Results}
\Cref{table:full-results} contain numbers corresponding to the graphs in Fig. 2 of the main text. 
Each table presents results for the different models under attacks of increasing strength ($\testRandDepthOne, \testRandDepthFive, \testGradDepthOne, \testGradDepthFive$). 
We note that, in all but one case, the adversarially trained model ($\advTrain{\testGradDepthOne}$), performs the best under attack.

\begin{table*}[t]
\caption{
Evaluation of our approach and baselines across four datasets.
Numbers in brackets are the difference in adversarial F1 compared to
the normally trained model (higher is better).
}
\label{table:full-results}
\subcaption{Results for the random 1-adversary ($\testRandDepthOne{}$).}
\resizebox{\textwidth}{!}
{
\begin{tabular}{cccccccccc}
\toprule
\multirow{3}{*}{Model} &
\multirow{3}{*}{Training} &
\multicolumn{2}{c}{c2s/java-small} &
\multicolumn{2}{c}{csn/java} &
\multicolumn{2}{c}{csn/python} & 
\multicolumn{2}{c}{sri/py150} \\
\cmidrule(lr){3-4} \cmidrule(lr){5-6} \cmidrule(lr){7-8} \cmidrule(lr){9-10}
& & 
$\testIdentity{}$ & $\testRandDepthOne{}$ &
$\testIdentity{}$ & $\testRandDepthOne{}$ &
$\testIdentity{}$ & $\testRandDepthOne{}$ &
$\testIdentity{}$ & $\testRandDepthOne{}$  \\
\midrule
\csvreader[
head to column names,
late after line=\\,
late after last line=\\\midrule
]{data/table-seq2seq.csv}{}%
{
  \ifthenelse{
    \equal{\modelName}{seq2seq}
  }{%
    \multirow{4}{*}{seq2seq}
  }{} &
  \trainingType & 
  \ctsJavaSmallId &
  \ctsJavaSmallQOneR &
  \csnJavaId &
  \csnJavaQOneR  &
  \csnPythonId &
  \csnPythonQOneR &
  \sriPyId &
  \sriPyQOneR
}%
\csvreader[
head to column names,
late after line=\\,
late after last line=\\\bottomrule
]{data/table-code2seq.csv}{}%
{
  \ifthenelse{
    \equal{\modelName}{code2seq}
  }{%
    \multirow{4}{*}{code2seq}
  }{} &
 \trainingType & 
 \ctsJavaSmallId &
 \ctsJavaSmallQOneR &
 \csnJavaId &
 \csnJavaQOneR &
 \csnPythonId &
 \csnPythonQOneR &
 \sriPyId &
 \sriPyQOneR
}%
\end{tabular}
}
\vspace{0.5em}
\subcaption{Results for the random 5-adversary ($\testRandDepthFive{}$).}
\resizebox{\textwidth}{!}
{
\begin{tabular}{cccccccccc}
\toprule
\multirow{3}{*}{Model} &
\multirow{3}{*}{Training} &
\multicolumn{2}{c}{c2s/java-small} &
\multicolumn{2}{c}{csn/java} &
\multicolumn{2}{c}{csn/python} & 
\multicolumn{2}{c}{sri/py150} \\
\cmidrule(lr){3-4} \cmidrule(lr){5-6} \cmidrule(lr){7-8} \cmidrule(lr){9-10}
& & 
$\testIdentity{}$ & $\testRandDepthFive{}$ &
$\testIdentity{}$ & $\testRandDepthFive{}$ &
$\testIdentity{}$ & $\testRandDepthFive{}$ &
$\testIdentity{}$ & $\testRandDepthFive{}$  \\
\midrule
\csvreader[
head to column names,
late after line=\\,
late after last line=\\\midrule
]{data/table-seq2seq.csv}{}%
{
  \ifthenelse{
    \equal{\modelName}{seq2seq}
  }{%
    \multirow{4}{*}{seq2seq}
  }{} &
  \trainingType & 
  \ctsJavaSmallId &
  \ctsJavaSmallQFiveR &
  \csnJavaId &
  \csnJavaQFiveR  &
  \csnPythonId &
  \csnPythonQFiveR &
  \sriPyId &
  \sriPyQFiveR
}%
\csvreader[
head to column names,
late after line=\\,
late after last line=\\\bottomrule
]{data/table-code2seq.csv}{}%
{
  \ifthenelse{
    \equal{\modelName}{code2seq}
  }{%
    \multirow{4}{*}{code2seq}
  }{} &
 \trainingType & 
 \ctsJavaSmallId &
 \ctsJavaSmallQFiveR &
 \csnJavaId &
 \csnJavaQFiveR &
 \csnPythonId &
 \csnPythonQFiveR &
 \sriPyId &
 \sriPyQFiveR
}%
\end{tabular}
}
\vspace{0.5em}
\subcaption{Results for the targeted 1-adversary ($\testGradDepthOne{}$).}
\resizebox{\textwidth}{!}
{
\begin{tabular}{cccccccccc}
\toprule
\multirow{3}{*}{Model} &
\multirow{3}{*}{Training} &
\multicolumn{2}{c}{c2s/java-small} &
\multicolumn{2}{c}{csn/java} &
\multicolumn{2}{c}{csn/python} & 
\multicolumn{2}{c}{sri/py150} \\
\cmidrule(lr){3-4} \cmidrule(lr){5-6} \cmidrule(lr){7-8} \cmidrule(lr){9-10}
& & 
$\testIdentity{}$ & $\testGradDepthOne{}$ &
$\testIdentity{}$ & $\testGradDepthOne{}$ &
$\testIdentity{}$ & $\testGradDepthOne{}$ &
$\testIdentity{}$ & $\testGradDepthOne{}$  \\
\midrule
\csvreader[
head to column names,
late after line=\\,
late after last line=\\\midrule
]{data/table-seq2seq.csv}{}%
{
  \ifthenelse{
    \equal{\modelName}{seq2seq}
  }{%
    \multirow{4}{*}{seq2seq}
  }{} &
  \trainingType & 
  \ctsJavaSmallId &
  \ctsJavaSmallQOneG &
  \csnJavaId &
  \csnJavaQOneG  &
  \csnPythonId &
  \csnPythonQOneG &
  \sriPyId &
  \sriPyQOneG
}%
\csvreader[
head to column names,
late after line=\\,
late after last line=\\\bottomrule
]{data/table-code2seq.csv}{}%
{
  \ifthenelse{
    \equal{\modelName}{code2seq}
  }{%
    \multirow{4}{*}{code2seq}
  }{} &
 \trainingType & 
 \ctsJavaSmallId &
 \ctsJavaSmallQOneG &
 \csnJavaId &
 \csnJavaQOneG &
 \csnPythonId &
 \csnPythonQOneG &
 \sriPyId &
 \sriPyQOneG
}%
\end{tabular}
}
\vspace{0.5em}
\subcaption{Results for the targeted 5-adversary ($\testGradDepthFive{}$).}
\resizebox{\textwidth}{!}
{
\begin{tabular}{cccccccccc}
\toprule
\multirow{3}{*}{Model} &
\multirow{3}{*}{Training} &
\multicolumn{2}{c}{c2s/java-small} &
\multicolumn{2}{c}{csn/java} &
\multicolumn{2}{c}{csn/python} & 
\multicolumn{2}{c}{sri/py150} \\
\cmidrule(lr){3-4} \cmidrule(lr){5-6} \cmidrule(lr){7-8} \cmidrule(lr){9-10}
& & 
$\testIdentity{}$ & $\testGradDepthFive{}$ &
$\testIdentity{}$ & $\testGradDepthFive{}$ &
$\testIdentity{}$ & $\testGradDepthFive{}$ &
$\testIdentity{}$ & $\testGradDepthFive{}$  \\
\midrule
\csvreader[
head to column names,
late after line=\\,
late after last line=\\\midrule
]{data/table-seq2seq.csv}{}%
{
  \ifthenelse{
    \equal{\modelName}{seq2seq}
  }{%
    \multirow{4}{*}{seq2seq}
  }{} &
  \trainingType & 
  \ctsJavaSmallId &
  \ctsJavaSmallQFiveG &
  \csnJavaId &
  \csnJavaQFiveG  &
  \csnPythonId &
  \csnPythonQFiveG &
  \sriPyId &
  \sriPyQFiveG
}%
\csvreader[
head to column names,
late after line=\\,
late after last line=\\\bottomrule
]{data/table-code2seq.csv}{}%
{
  \ifthenelse{
    \equal{\modelName}{code2seq}
  }{%
    \multirow{4}{*}{code2seq}
  }{} &
 \trainingType & 
 \ctsJavaSmallId &
 \ctsJavaSmallQFiveG &
 \csnJavaId &
 \csnJavaQFiveG &
 \csnPythonId &
 \csnPythonQFiveG &
 \sriPyId &
 \sriPyQFiveG
}%
\end{tabular}
}
\end{table*}
\clearpage



\end{document}